# Learning Random Fourier Features by Hybrid Constrained Optimization

**Jianqiao Wangni**  ZJNQHA@GMAIL.COM
**Jingwei Zhuo**  ZJW15@MAILS.TSINGHUA.EDU.CN
**Jun Zhu**  DCSZJ@MAIL.TSINGHUA.EDU.CN
Tsinghua University

## Abstract

The kernel embedding algorithm is an important component for adapting kernel methods to large datasets. Since the algorithm consumes a major computation cost in the testing phase, we propose a novel teacher-learner framework of learning computation-efficient kernel embeddings from specific data. In the framework, the high-precision embeddings (teacher) transfer the data information to the computation-efficient kernel embeddings (learner). We jointly select informative embedding functions and pursue an orthogonal transformation between two embeddings. We propose a novel approach of constrained variational expectation maximization (CVEM), where the alternate direction method of multiplier (ADMM) is applied over a nonconvex domain in the maximization step. We also propose two specific formulations based on the prevalent Random Fourier Feature (RFF), the masked and blocked version of Computation-Efficient RFF (CERF), by imposing a random binary mask or a block structure on the transformation matrix. By empirical studies of several applications on different real-world datasets, we demonstrate that the CERF significantly improves the performance of kernel methods upon the RFF, under certain arithmetic operation requirements, and suitable for structured matrix multiplication in Fastfood type algorithms.

## 1. Introduction

Nonlinear machine learning models are important tools for analyzing data. Among them, the kernel method (Schölkopf & Smola, 2002) is widely applied to the classifiers and dimensionality reduction techniques., such as the kernel support vector machine (KSVM), the kernel principal component analysis (KPCA), and the kernel canonical correlation analysis (KCCA). The kernel methods require calculating the similarity of data pairs in high-dimensional space while avoiding determining the exact representation in this space. However, the computational complexity of calculating kernel matrix increases quadratically with the dataset size, besides, storing a $N \times N$ Gram matrix of $N$ training samples puts heavy pressure on the computer memory. Consequently this approach scales poorly with modern large datasets, which continue to grow rapidly on the internet.

A representitive work on boosting the scalability of kernel methods is the *Random Fourier Feature method (RFF)* (Rahimi & Recht, 2009), whose computational complexity is $O(ND)$, significantly improved upon traditional methods. By building a probability distribution which is proportional to the Fourier transform of the kernel function, monte-carlo sampled feature functions will approximate the kernel function to any precision with an enough sampling size, with a theoretical guarantee. There are many variants, for examples, (Rick Chang et al., 2016) approximated sparse data using much lower dimensional embeddings; (Hamid et al., 2014) proposed Compact Random Features for approximating polynomial kernels with high accuracy; (Yang et al., 2014) proposed Random Laplace Features to adapt the RFF to the histogram type data defined in the semigroup of $\mathbb{R}^d_+$.

Despite the success and popularity of the aforementioned algorithms, the computation cost issue still needs further investigations. For instance, the RFF methods involve a large matrix-vector multiplication, whose cost is proportional to the dimension of embeddings and data. There are some works improving the efficiency of the random fourier features from a computing method perspective, for an example, the Fastfood algorithm (Le et al., 2013) adopted a combination of structured matrices and diagonal Gaussian matrices to approximate the properties of dense Gaussian random matrices, requiring $O(n \log d)$ time and $O(n)$ storage to compute $n$ non-linear basis functions in $d$ dimen-sions. Many algorithms based on the idea are later proposed in (Feng et al., 2015)(Choromanski & Sindhwani, 2016)(Yu et al., 2015)(Rick Chang et al., 2016)(Hamid et al., 2014).

In this paper, we are focusing on training kernel embeddings from data, aiming to improve their performance under a computational cost budget in the testing phase. The idea of training kernels from data is well studied in previous literature, for examples, the Multiple Kernel Learning (MKL) (Bach et al., 2004) and the Local Deep Kernel (Jose et al., 2013) learn a non-negative linear combination of fixed kernels from a flexible class. Our primary consideration is, since directly learning kernel embedding functions from a large Gram matrix is both computational and memory consuming, we consider a teacher-learner framework, to be specific, for each data sample, its high-cost embedding (teacher) transfer the information to its low-cost one (learner). This approach avoids the explicit calculation of Gram matrices and is easily developed into stochastic variants, therefore is scalable for training. Based on the consideration, we propose a novel and generalized kernel learning framework, the *Computation-Efficient Random Fourier Features (CERF)*. We also propose two specific formulations, the first one is to place a sparse mask matrix on the CERF transformation matrix, and the second one is to explicitly make the CERF in block-wise form, whose each block is accelerated by the Fastfood algorithm.

### 1.1. Background

The *Random Fourier Feature (RFF)* (Rahimi & Recht, 2009) is a method of generating low-dimensional embeddings of shift invariant kernels, which only relies on the difference between two data samples, $\mathcal{K}(x, y) = \mathcal{K}(x - y, 0)$, where $x, y \in \mathbb{R}^D$.

**Definition 1** *The RFF method transforms the data to their kernel embeddings $\phi : \mathbb{R}^D \to \mathbb{R}^K$ by sampling $K$ transformation vectors $\omega \in \mathbb{R}^D$ i.i.d from $p(\omega)$,*

$$\phi(x) = \frac{1}{\sqrt{K}}[\cos(\omega_1^\top x + b_1), \cdots, \cos(\omega_K^\top x + b_K)]^\top. \quad (1)$$

*where the parameter distribution is obtained by normalizing the Fourier transformation of the kernel function,*

$$p(\omega) \propto \int \mathcal{K}(x, 0) e^{-i\omega^\top x} dx. \quad (2)$$

*and the bias $b \in \mathbb{R}$ is a random variable following a uniform distribution $[0, 2\pi]$.*

In addition, the parameter distribution of RBF kernel and the Cauchy kernel is the Gaussian distribution and Laplace distribution, respectively.

The *Beta-Bernoulli Process (BBP)* (Paisley & Carin, 2009) is an important building block for our model later introduced, it is a prior on binary matrices which represents the presence of features for each data sample by rows, and each element of a row represents whether one feature is activated for this sample.

**Definition 2** *Suppose a binary feature presence matrix $Z \in \{0, 1\}^{N \times K}$ of $N$ samples and $K$ features, is sampled from a Beta-Bernoulli Process, the probability of the $k$-th feature activated by each sample is defined by $\pi_k$, and the probability of $Z \sim BBP(\gamma)$ is determined by*

$$P(Z|\gamma) = \prod_k \prod_n P(z_{nk}|\pi_k) = \prod_k \pi_k^{m_k}(1 - \pi_k)^{N-m_k}, \quad (3)$$

*where $m_k = \sum_{n=1}^N z_{nk}$ sums up all the data samples which are activated by feature $k$. The prior on $\pi$ is the Beta distribution,*

$$\pi_k \sim Beta(\gamma/K, 1), \quad z_{nk} \sim Bernoulli(\pi_k). \quad (4)$$

## 2. Computation-Efficient Kernel Embeddings

A straightforward method for learning a computational low-cost kernel embedding $\varphi$, is to explicitly force it to approximate the shift-invariant kernel $\mathcal{K}$ defined on $\mathbb{R}^D \times \mathbb{R}^D$,

$$\min_{\varphi : \mathbb{R}^D \to \mathbb{R}^K} \mathbb{E}_{x, y \sim \mathcal{D}} \left[ \left( \mathcal{K}(x, y) - c^2 \varphi(x)^\top \varphi(y) \right)^2 \right], \quad (5)$$

where $\mathcal{D}$ is the data distribution, and $c$ is a constant. But this method is not scalable due to the $O(N^2)$ computation and storing of the large Gram matrix, like traditional kernel methods. Instead, we will seek the best low-cost kernel embeddings $\varphi(x) \in \mathbb{R}^K$ (learner) to approximates the high-cost kernel embeddings $\phi(x) \in \mathbb{R}^K$ (teacher). In our framework, we do not restrict $\phi(x)$ to be any specific kind, it can be obtained by Nystrom method (Drineas & Mahoney, 2005) or Kernel PCA (Schölkopf et al., 1998). We study a specific implementation, that $\phi(x)$ is obtained from the RFF embedding. The approximation requirement is transferred to be on the embedding function

$$\phi(x)^\top \phi(y) \approx c^2 \varphi(x)^\top \varphi(y), \quad \phi(x)^\top \phi(y) \approx \mathcal{K}(x, y), \quad (6)$$

This problem has improvement over Eq.(5), since it only consumes $O(ND)$ computations and memory. To continue simplifying the problem, we take a heuristic approach of finding a linear transformation matrix $W$ that connects the two kinds of kernel embeddings,

$$\phi(x) \approx c W \varphi(x), \quad W \in \Omega, \quad (7)$$
$$\text{where} \quad \Omega = \{W \in \mathbb{R}^{K \times K} | W^\top W = I_K\}, \quad (8)$$

and $I_K \in \mathbb{R}^{K \times K}$ is an identity matrix. As a result, the orthogonality further implies that

$$\phi(x)^\top \phi(y) \approx c^2 \varphi(x)^\top W^\top W \varphi(y) = c^2 \varphi(x)^\top \varphi(y), \quad (9)$$

which leads to our main target. We define a likelihood function $p$ for modeling the approximation between the two kinds of embeddings of $x$,

$$p(\phi(x), \varphi(x)|z, W) = \exp(-\frac{||\phi(x) - cW(z \odot \varphi(x))||^2}{2\sigma^2}).$$

where $\sigma$ is a parameter. A larger probability implies a higher approximation precision. We denote $\phi_n = \phi(x_n)$, $\varphi_n = \varphi(x_n)$, $\Phi$, $\Psi$ and $Z$ as matrices which column-wisely stacking $\phi_n$, $\varphi_n$ and $z_n$, respectively, and $Q = \{Z, \pi\}$. By the Bayes rule, the posterior probability of $Q$ is

$$p(\Phi, \Psi, Q|W) \propto p_0(Q) \prod_{n=1}^{N} p(\phi_n, \varphi_n|z_n, W). \quad (10)$$

Our target is to maximize the likelihood of $W$ by marginalizing over the latent variables $Q$,

$$\log p(\Phi, \Psi|W) = \log \int p(\Phi, \Psi, Q|W)dQ. \quad (11)$$

For optimizing the probability in Eq.(11), we use both variational inference and convex optimization methods, and apply a convex relaxation technique to deal with the orthogonality constraint. The details are as follows.

### 2.1. Constrained Variational Expectation Maximization

To avoid the explicit marginalization in Eq.(11), we heuristically optimize a variational lower bound with respect to the latent variables $Q$, by applying the Jensen inequality

$$\log p(\Phi, \Psi|W) = \log \int q(Q)\frac{p(\Phi, \Psi, Q|W)}{q(Q)}dQ \quad (12)$$

$$\geq \int q(Q)(\log p(\Phi, \Psi, Q|W) - \log q(Q))dQ \quad (13)$$

$$= \mathbb{H}(q(Q)) + \mathbb{E}_q[\log p(\Phi, \Psi, Q|W)], \quad (14)$$

Denoting $\mathbb{H}(q(Q))$ is the entropy of the variational distribution. The function

$$\mathcal{L}(q(Q), W) = \mathbb{H}(q) + \mathbb{E}_q[\log p(\Phi, \Psi, Q|W)] \quad (15)$$

is the result of the log posterior $\log p(\Phi, \Psi, Q|W)$ substracts the KL divergence $KL(p||q)$ between the variational distribution $q(Q)$ and the true posterior $p(\Phi, \Psi, Q|W)$. So, maximizing over $\mathcal{L}(q(Q), W)$ acts like minimizing the term $KL(p||q)$.

The overall algorithm can be described by the expectation maximization (EM) framework, in which the objective function $\mathcal{L}(q(Q), W)$ with respect to $q(Q)$ and $W$ is iteratively optimized. In the $l$-th stage, we apply the following steps to find a local optima,

E-step: $\hat{q}^l(Q) = \arg\max_{q(Q)} \mathcal{L}(q(Q), \hat{W}^{l-1}), \quad (16)$

M-step: $\hat{W}^l = \arg\max_{W \in \Omega} \mathcal{L}(\hat{q}^l(Q), W). \quad (17)$

To make the E-step solvable, we make a general mean-field assumption that the variational distribution is factorizable, and each individual component of the distribution owns an parameter,

$$q(Q) = q(\pi)q(Z), \quad q(\pi_k) = Beta(\tau_{k1}, \tau_{k2}), \quad (18)$$
$$q(z_{nk}) = Bernoulli(\nu_{nk}). \quad (19)$$

For the E-step in the $l$-th stage, we follow the standard procedure of coordinate descent, then optimize each variational parameter, meanwhile fixing the others. This will produce a sequence of updated variables with a convergence guarantee. We will use $q^t(Q)$ denoting the variational distribution in the $t$-th step, and use $q^{t+1}(Q)$ denoting the one in the next step accordingly. Generally speaking, the update equation for the $i$-th parameter $Q_i$ in the variational inference framework is as following,

$$q^{t+1}(Q_i) \propto \exp(\mathbb{E}_{q^t(Q_{-i})}[\ln p(\Phi, \Psi, Q_i|W)]), \quad (20)$$

where $\mathbb{E}_{q(Q_{-i})}$ means taking expectation with respect to all parameters other than $Q_i$.

Now we analyse the update equations for each variational parameter under the principle in Eq.(20). First, we denote $\xi_{nk} \in \mathbb{R}^K$ as the reconstruction error vector using all embedding functions only without the $k$-th one,

$$\xi_{nk} = \phi_n - cW^\top(\varphi_n \odot z_n) + c\varphi_{nk}z_{nk}W_k^\top, \quad (21)$$

By the Bayes rule, we have

$$p(\phi_n, \varphi_n, z_{nk}|W) \propto p(\phi_n, \varphi_n, |z_{nk}, W)p_0(z_{nk}|\pi_k)$$
$$= \pi_k \exp(-\frac{1}{2\sigma^2}||\xi_{nk}||_2^2).$$

By simplifying the equation above, and following Eq.(20), the update rule for $q(z_{nk})$ is

$$q^{t+1}(z_{nk} = 1) \propto \exp(\mathbb{E}_{q^t}[\ln(\pi_k) - \frac{\Delta_{nk}}{2\sigma^2}]), \quad (22)$$
$$q^{t+1}(z_{nk} = 0) \propto \exp(\mathbb{E}_{q^t}[\ln(1 - \pi_k)]), \quad (23)$$

where $\mathbb{E}_{q^t}[z_k] = \nu_{nk}^t$, and

$$\Delta_{nk} = c^2\varphi_{nk}^2 W_k^\top W_k - 2c\varphi_{nk}W_k^\top \xi_{nk}. \quad (24)$$

By the property of beta distribution,

$$\exp(\mathbb{E}_{q^t}[\ln(\pi_k)]) = \psi(\tau_{k1}^t) - \psi(\tau_{k1}^t + \tau_{k2}^t), \quad (25)$$
$$\exp(\mathbb{E}_{q^t}[\ln(1 - \pi_k)]) = \psi(\tau_{k2}^t) - \psi(\tau_{k1}^t + \tau_{k2}^t).$$

where $\psi(\cdot)$ is a dig-gamma function. Substituting Eq.(25) to Eq.(22), then $\nu_{nk}$ is updated by

$$\nu_{nk}^{t+1} = q^{t+1}(z_{nk} = 1)/(q^{t+1}(z_{nk} = 1) + q^{t+1}(z_{nk} = 0)).$$

By Eq.(20), the update for $q(\pi_k)$ is

$$q^{t+1}(\pi_k) \propto p_0(\pi_k) \prod_{n=1} \exp\left([\nu_{nk}^t \ln \pi_k + (1-\nu_{nk}^t)\ln(1-\pi_k)]\right)$$
$$\propto \exp\left(c_0 + \sum_n \nu_{nk}^t\right),$$

and the simplified update rule is

$$\tau_{k1}^{t+1} = \sum_n \nu_{nk}^t + \gamma/K, \quad \tau_{k2}^{t+1} = \sum_n (1-\nu_{nk}^t) + 1. \quad (26)$$

By assigning $W = \hat{W}^{l-1}$ in the beginning, and assigning $\hat{q}^l = q^t(Q)$ in the end, we finish the E-step in the $l$-th stage.

### 2.2. M-step with ADMM

The stage-wise M-step with fixed $\hat{q}^l(Z)$ is by the obtained by the minimization over the following function

$$W^l = \arg\min_{W \in \Omega} ||\Phi - cW(\mathbb{E}_{\hat{q}^l(Q)}[Z] \odot \Psi)||_F^2, \quad (27)$$

The optimization on the orthogonality constrained set $\Omega$ in Eq.(27) is nonconvex, therefore it is sensitive to initialization and very likely converges to a poor local optima.

This drawback was also met in the stochastic solver of principal component analysis, where the components are required to be orthogonal. In some 3D vision problems, the rotation matrix of viewpoint which needs estimation is also orthogonal. The Procrustes method, which projects the estimated matrix to the orthogonal group via singular value decomposition (SVD), gives a simple solution but not a closed-form one. We are inspired by the 3D vision work (Zhou et al., 2015), which replaces the orthogonality constraint by a convex counterpart.

We cite a property from (Journée et al., 2010), Section 3.4, that the unit spectral-norm ball $conv(\Omega) = \{W \in \mathbb{R}^{K \times K} \mid ||W||_2 \le 1\}$ is tightest convex outer hull of the orthogonality constrained set $\Omega = \{W^\top W = I_K\}$. Building on this, we put a spectral norm regularization substituting the hard constraint,

$$\arg\min_W ||\Phi - cW(\mathbb{E}_{\hat{q}^l(Q)}[Z] \odot \Psi)||_F^2 + \alpha ||W||_2. \quad (28)$$

where $\alpha$ is an adaptive parameter. Note that the spectral norm equals to the largest singular value of a matrix. We optimize the problem in Eq.(28) using the alternate direction method of multiplier method (ADMM), which reduces a problem with complex regularization to several smaller ones with independent regularization. We introduce an auxiliary variables $V$, a dual variable $U$ and a stepsize parameter $\mu$ to form the augmented Lagrangian of Eq.(28) as

$$L_\mu(W, V, U) = \frac{1}{2}||\Phi - cV(\mathbb{E}_{\hat{q}^l(Q)}[Z] \odot \Psi)||_F^2 + \alpha ||W||_2$$
$$+ U^\top(W - V) + \frac{\mu}{2}||W - V||^2.$$

Then, the ADMM updates each variable $W$, $V$ and $U$ iteratively while fixing other variables until convergence. Following the proof provided by (Boyd et al., 2011), this updating strategy produces a variable sequence that converges to the optima of the problem in Eq.(28). Since the function $L_\mu(W, V, U)$ is quadratic w.r.t. $V$, so there exists a closed-form update. The update for $V$ and $U$ is in closed-form,

$$V^{t+1} = (\Phi \bar{\Psi}^\top + \mu W^t + U^t)(\bar{\Psi}\bar{\Psi}^\top + \mu I)^{-1}, \quad (29)$$
$$U^{t+1} = U^t + \mu(W^t - V^{t+1}). \quad (30)$$

where

$$\bar{\Psi} = \mathbb{E}_{\hat{q}^l}[Z] \odot \Psi. \quad (31)$$

The update for $W$ is a little troublesome, we rewrite Eq.(28) as a proximal problem w.r.t. $W$,

$$\arg\min_W \frac{\mu}{2}||W - V^{t+1} + \frac{1}{\mu}U^{t+1}||^2 + \alpha ||W||_2, \quad (32)$$

and denote the solution by $prox_\beta(A)$, where $A = V^{t+1} - U^{t+1}/\mu$ and $\beta = 2\alpha/\mu$. Suppose that $A$ has single-value decomposition as $A = R\,\text{diag}[\sigma_A]P^\top$, where $\sigma_A$ stacks singular values in descending order. By a property of spectral functions in [(Parikh et al., 2014), Section 6.7.2], then

$$prox_\beta(A) = R\,\text{diag}[\sigma_A - \beta\mathcal{P}_1(\sigma_A/\beta, 1)]P^\top, \quad (33)$$

where $P_1(v, a)$ is the Euclidean projection onto the $L_1$ ball of radius $a$, $P_1(v, a) = \arg\min_{x:||x||_1 \le a} \frac{1}{2}||x - v||^2$, $x, v \in \mathbb{R}^K$. Following (Liu & Ye, 2009), for the case $||v||_1 \le a$, $P_1(v, a)$ is directly assigned by $v$; otherwise, $||v||_1 \ge a$, then we add a Lagrangian variable $\lambda$ for the regularization $||v||_1 \le a$ and rewrite it as

$$P_1(v, a) = \frac{1}{2}||x - v||^2 + \lambda(||x||_1 - a), \quad \lambda > 0. \quad (34)$$

After obtaining the optimal dual variable $\lambda^\star$ by solving $f(\lambda) = \sum_{i=1}^D \max(|v_i| - \lambda, 0) - z = 0$, then $P_1(v, a) = \text{sign}(v_i)\max(|v_i| - \lambda^\star, 0)$. Combining the updates above, and assigning $\hat{W}^l = W^t$ in the end, we finish the M-step in the $l$-th stage. After the algorithm convergences, we orthogonalize $W$ by projecting it to the constrained set $\Omega$ and add several more stages while fixing $W$ to encourage the inner product approximation of CERF to the target.

## 3. Applications in Random Fourier Features

We will introduce two examples of the nonlinear mapping functions $\varphi(x)$, they are based on the aforementioned random fourier features.

We rewrite $\varphi(x)$ in a element-wise form, $[\varphi_1(x), \cdots, \varphi_K(x)]^\top$. As we can see from the formulation that each $\varphi_k(x)$ is densely connected to all dimensions of $x$, so we use this as a starting point for improvement. Inspired by sparse modelling literature, we introduce $\epsilon_k \in \{0,1\}^D$ as binary mask vectors, each indicating whether a data element is used for the $k$-th feature or not, and $\odot$ is the element-wise product operator. Since only the masked elements need computation in testing phase, the consequent speed is accelerated. Here we sample $\epsilon$ from the BBP described above, which is fixed during training.

**Proposition 3** *By sampling $\omega_k$ and $b_k$ with RFF, The Masked Random Fourier Feature (masked CERF) function $\varphi(x)$ is defined as*

$$\varphi_k(x) = \cos(\rho(x \odot \epsilon_k)^\top \omega_k + b_k), \tag{35}$$
$$\text{where } \epsilon \sim BBP(\gamma), \epsilon_k \in \{0,1\}^D.$$

For the kernels whose RFF are drew from irregular distributions, we can learn a CERF embedding which is easily accelerated by the Fastfood algorithm (Le et al., 2013).

**Definition 4** *For the RBF kernel, denoting the Hadamard matrix as $H$, a permutation matrix $Q$, a binary scaling matrix as $B$, a diagonal Gaussian matrix $G$ and a scaling matrix $S$, then the production*

$$V = \frac{1}{\sigma} SHGQHB \tag{36}$$

*follows the Gaussian distribution, but only costs $K \log(D)$ computations.*

Since a high-precision embedding typically has a dimension $K$ higher than original input $D$, we propose to use CERF functions which are formed into $K/D$ blocks of $D$ functions, the functions in each block follows a common distribution, and is therefore efficient to calculate by Fastfood or other structured matrices multiplication algorithms (Sindhwani et al., 2015) (Gu et al., 2016). With the blocked CERF, the computation is reduced to $K \log D$ from $KD$.

**Proposition 5** *The $K = JD$ dimensional Blocked Random Fourier Features (Blocked CERF) of $D$ dimensional data is defined as*

$$\varphi(x) = \cos(\Xi^\top x + b), \quad \text{where } \Xi \in \mathbb{R}^{D \times JD},$$
$$\Xi_{1:D, jD:(j+1)D} = \frac{1}{\sigma} SHG_{k,j} QHB, \quad \forall j \in [J],$$

## 4. Experiments

We conducted our experiments on various real world datasets. We use the NUS-WIDE dataset (Chua et al., 2009), which is collected from Flicker for image retrieval tasks, it consists of 27K images, all with tags. We use four kinds of features, including 64-D color histogram *(CH)*, 144-D color correlogram *(CORR)*, 73-D edge direction histogram *(EDH)*, 225-D block-wise color moments *(CM)*. We randomly choose 500 images from the widely used *MNIST* dataset for training, and another 10,000 images for testing. Different from MNIST, the *digits-raw* dataset consists of 5,000 handwritten digit images in unit8 type and $20 \times 20$ resolution. For this dataset, we also build the *digits-hog* dataset by extracting 64 dimensional Histogram-Of-Gradient features. The latin character dataset, *Letters*, is designed for structured prediction models (Roller, 2004), we use the orderless version and perform PCA on the original data to compress to 70 dimensions. The training set and the testing set consists of 5,375 and 46,777 images respectively. The *Olivatte* face dataset consists of 200 face images. The *ORL* dataset consists of 400 face images. We also take medical datasets *Pima* and *Diabete* on diabetic patients, and the *Covtype* on forest cover type, from UCI machine learning repository.

We conduct our experiment by Python programs. We also use the *Scikit-Learn* machine learning package in implementation. We set the maximum stages of the variational EM framework to be 20, since the algorithm already converges well enough by this much. All the CERF features for the testing data share one feature selector $z$, which is determined by the final values of the variational distribution $\hat{q}^l(\pi)$. We also set the nonzero item density of mask vectors $\epsilon_k$ to be $40\%$ for all $k$, and the nonzero item density of the testing CERF embeddings to be $20\%$ or $40\%$. So the computation cost is reduced to $8\%$ and $16\%$ respectively. The best kernel parameter $\kappa^2$ is grid-searched within a large range of $\{10^{+3}, \cdots, 10^{-6}\}$ for the best outcome, and each parameter is checked by a 5-folds cross validation. It is also worthwhile to mention that, although a sparse formulation of CERF will enlarge its feature dimension and leads to more computations when applying classifiers or PCA, this enlarged computation complexities are $O(K|\mathcal{Y}|)$ and $O(KJ)$ respectively, considerably negligible comparing to the major cost $O(KD)$, where $\mathcal{Y}$ is the domain of labels, and $J$ is the reduced dimensions.

*Table 1.* Dataset information

| dataset | CH | CORR | EDH | CM |
|---|---|---|---|---|
| number | 27K | 27K | 27K | 27K |
| dimension | 64 | 144 | 73 | 225 |
| dataset | MNIST | Letter | Hog | Raw |
| number | 10,000 | 5,375 | 5,000 | 5,000 |
| dimension | 784 | 70 | 64 | 400 |
| dataset | Olivatte | ORL | Pima | Diabete |
| number | 200 | 400 | 768 | 1,151 |
| dimension | 4,096 | 1,024 | 8 | 20 |

Table 2. Normalized kernel approximation error with blocked CERF and blocked RFF (same computation complexity).

| data | $\mathcal{P}_1$ | param | $K'/K = 0.4$ Cerf | RFF | $K'/K = 0.2$ Cerf | RFF |
|---|---|---|---|---|---|---|
| raw | Gauss | 2.5e-5 | 88.7 | 84.1 | 114 | 141 |
| raw | Gauss | 5e-5 | 133 | 355 | 264 | 255 |
| raw | cauchy | 5e-5 | 119 | 248 | 156 | 232 |
| raw | cauchy | 1e-4 | 150 | 361 | 209 | 445 |
| hog | Gauss | 1e-3 | 104 | 120 | 365 | 647 |
| hog | cauchy | 1e-3 | 108 | 228 | 141 | 306 |

### 4.1. Kernel Approximation

We test the performance of approximating irregular kernel embeddings using the Block CERF, further accelerated by Fastfood. We use Digits-Raw AND Digits-Hog as testing datasets. The dimension of dataset is reduced to 16 by principal component analysis, before calculating the CERF or RFF. The parameter $\omega_k$ of our trained CERF function $\varphi(x) = \cos((\epsilon_k \odot x)^\top \omega_k + b_k)$ is sampled from a mixture of gaussian distribution, where each distribution is centered at a point drew from a main distribution $\mathcal{P}_0$. The parameter $\omega_k \in \mathbb{R}^{D \times K}$ of the target RFF $\phi(x) = \cos(x^\top \omega'_k + b_k)$, is the element-wise production of two sampled vectors, $\omega_k^{(1)} \sim \mathcal{P}_1$ and $\omega_k^{(2)} \sim \mathcal{P}_2$, as $\omega_k = \omega_k^{(1)} \odot \omega_k^{(2)}$, and we choose $\mathcal{P}_1$ to be the Gaussian or Cauchy distribution, and $\mathcal{P}_2$ to be the Gaussian distribution. We make the $\mathcal{P}_0$ to have same formulation and a same parameter with $\mathcal{P}_1$. We compared the results of trained CERF versus random sampled RFF sampled from $\mathcal{P}_0$ of the same dimension of CERF, which has the same computation complexity for the Fastfood acceleration. We set $K = 16 \times 20$ for all experiments, implying the CERF is divided into 20 blocks. We denote the CERF dimension as $K'$. Both RFF and CERF are selected using the aforementioned framework and the CVEM algorithm. In Table.(2), we report the error of calculated kernel matrix with the ground truth one, as $\frac{1}{N^2}\sum_{n=1}^{N}\sum_{m=1}^{N}|\phi(x_n)^\top\phi(x_m) - \varphi(x_n)^\top\varphi(x_m)|$, and the statistics are normalized. By comparing the results, we see that CERF can approximate an irregular kernel more accurate than RFF of the same computational cost.

### 4.2. Classification and Regression

We use the CERF embedding to perform Kernelized Support Vector machines (KSVM) for classification on multiple datasets described above. We use the LIBSVM implementation by (Chang & Lin, 2011), where the regularization parameter is grid searched for the best performance of RFF, and we directly transfer the parameter to the setting of CERF, so the comparison is fair. We divide each dataset into 5 folds and perform the experiments 5 times under each setting, using 4 folds for training and the left fold for testing, respectively. The computational cost is measured by **MAC** (Multiply And Accumulate) operations. We report the averaged accuracy in Table (5), which shows that the prediction accuracy with CERF is much higher than the corresponding RFF opponent. We also compare CERF with other methods for classification tasks. Our competitors include RKS (Rahimi & Recht, 2009), MKL (Bach et al., 2004), AlaC (Yang et al., 2015) and BaNK (Yang et al., 2015). The feature dimension is set to be 768 for fair comparisons. In Table (4), we see that the CERF achieves the highest accuracy in kernel methods.

Table 3. Classification Accuracy (%) noted as **Acc** on various datasets. The CERF and RFF are with the same **MAC** operations in each column.

| MAC | Acc | Acc | MAC | Acc | Acc |
|---|---|---|---|---|---|
| MNIST | CERF | RFF | Letter | CERF | RFF |
| 8K | 60.0 | 49.3 | 700 | 51.5 | 30.8 |
| 9K | 64.0 | 51.5 | 840 | 53.6 | 41.5 |
| 11K | 65.2 | 52.5 | 980 | 57.4 | 39.5 |
| 12K | 69.5 | 59.8 | 1.1K | 58.5 | 45.1 |
| 14K | 67.0 | 53.1 | 1.2K | 62.4 | 42.8 |
| 16K | 67.1 | 58.0 | 1.4K | 63.0 | 44.3 |
| RAW | Cerf | RFF | HOG | Cerf | RFF |
| 4K | 67.1 | 46.6 | 640 | 74.8 | 42.8 |
| 4.8K | 72.1 | 53.6 | 768 | 80.0 | 50.8 |
| 5.6K | 71.0 | 52.4 | 896 | 81.4 | 51.9 |
| 6.4K | 75.5 | 59.2 | 1.0K | 84.0 | 53.7 |
| 7.2K | 77.8 | 62.8 | 1.1K | 87.2 | 62.9 |
| 8K | 78.3 | 66.7 | 1.3K | 87.1 | 60.6 |
| Olivatte | Cerf | RFF | ORL | Cerf | RFF |
| 20K | 90.5 | 71.3 | 24K | 91.8 | 71.0 |
| 28K | 93.5 | 82.8 | 32K | 95.8 | 86.0 |
| 36K | 91.0 | 88.3 | 41K | 93.8 | 89.3 |
| 41k | 47.0 | 40.3 | 49k | 46.7 | 43.0 |
| 57K | 51.0 | 40.0 | 65K | 56.5 | 53.8 |
| 74K | 56.5 | 56.0 | 82K | 59.3 | 57.0 |

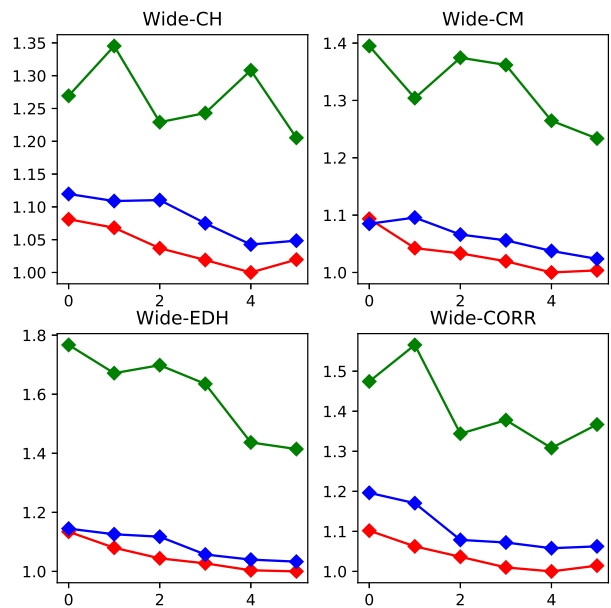

Figure 1. Normalized reconstruction error of randomized auto-encoder using different features of NUS-WIDE.

lution that only consumes $O(K^2 N)$ computations. Since the calculation of RFF in RKPCA is also time-consuming, we can use CERF to substitute the representation. We make the singular value decomposition of data as $\varphi(x) = R \operatorname{diag}[\sigma] P^\top$, where $\sigma$ stacks the singular values in descending order, then compute the linear embedding by $\varphi(x) P_J \in \mathbb{R}^{K \times J}$, using the top-$J$ singular vectors. In the following experiments, the dimension of CERF averagely spans in the range from 40 to 160, and the density of mask matrix $\epsilon$ is $20\%$ for NUS-wide and $40\%$ for other datasets, and the density of $z$ are all set to be $40\%$. By multiplication with feature selector density, the computation cost is reduced to $8\%$ and $16\%$ respectively. The corresponding RFF opponents are set to have equalized computation costs.

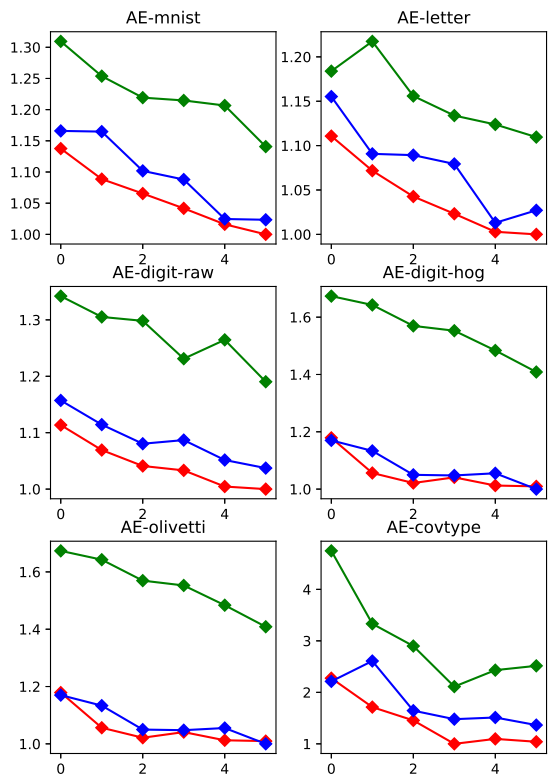

*Figure 2.* Reconstruction error of autoencoders. X-axis: the embedding dimension $K$ ascending from 40 to 160 with fixed stepsize. Y-axis: the normalized errors. Green: standard RFF, Blue CERF without training, Red: trained CERF.

*Table 4.* Classification Error (%) on LIBSVM datasets.

|  | RKS | MKL | AlaC | BaNK | CERF |
|---|---|---|---|---|---|
| pima | 33.2 | 44.55 | 25.92 | 26.3 | 23.6 |
| diabetic | 33.12 | 30.51 | 42.63 | 27.9 | 25.2 |

We will apply the kernel principal component analysis (KPCA) (Schölkopf et al., 1998) in the following experiments, following the randomized kernel PCA (RKPCA) (Lopez-Paz et al., 2014), which performs the principal component analysis (PCA) on the RFF embedding of original data, which leads to a scalable and randomized so-

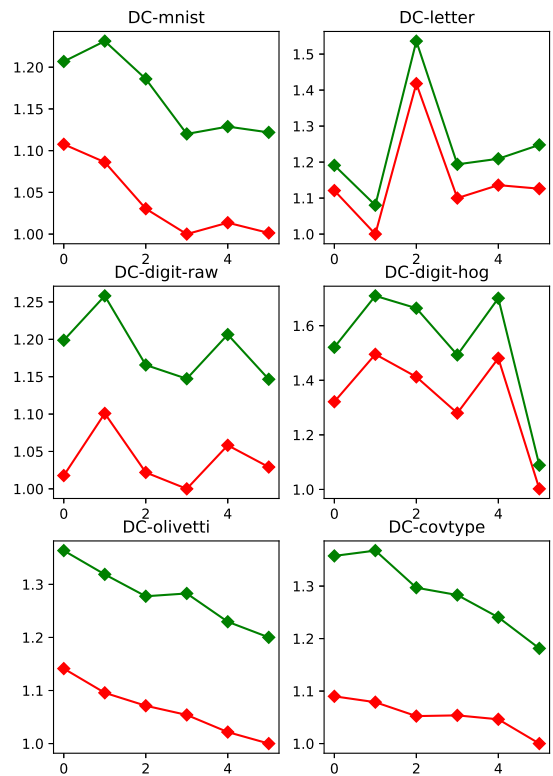

*Figure 3.* Normalized data completion error. Green: standard RFF; Red: trained CERF; X-axis: the embedding dimension $K$ ascending from 40 to 160.

We use CERF based RKPCA in the application of randomized autoencoders, to reconstruct the original data from kernel space. By performing the RKPCA, we obtain the even lower dimensional kernel embedding $\varphi(x)^\top P_J \in \mathbb{R}^J$. The reduced dimensions of all experiments are set to be $J = 15$ for fair comparisons. We reconstruct the original data by $D$ independent ridge regressors, from the highly compressed embeddings. Figure (1) shows the normalized reconstruction error on multiple datasets. The results undoubtedly show that CERF outperforms traditional RFF under all pa-rameter settings. Furthermore, it shows that the CERF outperforms by a larger margin when the computation requirement is higher, since the cost-reducing technique is more critical under such a setting. The results further prove the efficiency of CERF even with a sequential dimension reduction.

We also use a combination of CERF and RKPCA for data completion, where we randomly blocked out 20% dimensions of the original data. We denote the incomplete data as $x \odot \epsilon_c$, where $\epsilon_c \in \{0,1\}^D$ and $\sum_d (\epsilon_{cd})/D = 0.8$. By computing the CERF embedding from the incomplete data $x \odot \epsilon_c$ and reducing to $J = 15$ dimensions by PCA, then we use the $J$ dimensional embedding to complete the missing data by $D' = \sum_d (1 - \epsilon_{cd})$ independent ridge regressors. Figure (3) shows the data completion error using CERF and RFF respectively, the results are also very discriminative across different embeddings.

Table 5. Reconstruction Errors by kernelized regression, with structured tranformation $\varphi$ of groupsize $2^e$, spanning from 1 to 16, with different scale of MAC operations.

| Data | MAC | 1 | 2 | 4 | 8 | 16 |
|---|---|---|---|---|---|---|
| RAW | 4K | 1730 | 1733 | 1786 | 1713 | 1717 |
|  | 4.8K | 1676 | 1650 | 1649 | 1635 | 1675 |
|  | 5.6K | 1620 | 1628 | 1602 | 1652 | 1636 |
|  | 6.4K | 1608 | 1592 | 1622 | 1573 | 1615 |
|  | 7.2K | 1563 | 1580 | 1565 | 1571 | 1607 |
|  | 8K | 1560 | 1550 | 1579 | 1572 | 1576 |
| HOG | 640 | 1076 | 1054 | 1071 | 1084 | 1127 |
|  | 768 | 956.7 | 1006 | 1046 | 1034 | 1061 |
|  | 896 | 939.3 | 943.1 | 984.7 | 997.9 | 1004 |
|  | 1.0K | 946.9 | 939.7 | 940.7 | 979.2 | 991.6 |
|  | 1.1K | 920.7 | 908.4 | 926.1 | 935.5 | 959.9 |
|  | 1.3K | 920.4 | 907.3 | 931.7 | 936.5 | 971.9 |
| Olivatte | 41k | 7186 | 7184 | 7308 | 7130 | 7210 |
|  | 49k | 7076 | 7027 | 6958 | 7103 | 7039 |
|  | 57K | 6705 | 6883 | 6558 | 6566 | 6498 |
|  | 65K | 6541 | 6743 | 6569 | 6813 | 6507 |
|  | 74K | 6392 | 6593 | 6647 | 6452 | 6458 |
|  | 82K | 6683 | 6527 | 6358 | 6515 | 6317 |

The results of these three tasks are reasonable, by two major reasons. First, each CERF function is only related to a small part of the original data, this modification clearly saves the computation cost of the RFF to a great degree, while the introduction of the mask matrix $\epsilon$ does not effect the data variance, by the property of BBP. So each CERF function can well approximate the 'teacher', i.e. the RFF function. In addition, by the low-cost nature of the CERF, we can sample more CERF function than RFF under a same cost budget, this gives a typical reason of why even untrained CERF is better than RFF. Second, since the CERF features are specifically selected from a large dictionary of RFF features by their importance to a predefined data distribution, the consequent performance (classification, data reconstruction, data completion) should be better when evaluating on the same data distribution.

### 4.3. Hardware Friendly Adaptation

Here we discuss the deployment on Single Instructions Multiple Data (SIMD) instruction set supported Modern CPUs, like *SSE* on Intel CPUs or *Neon* on ARM CPUs. The blocked CERF is already hardware-friendly. For the masked CERF, By making a little modification that each element $\epsilon_i$, $1 \leq i \leq d$ of masking vectors $\epsilon \in \{0,1\}^d$ indicates the activation of sequential $2^e$ elements of $x$, where $d = 2^{-e}D$ and $e \in Z^+$. This will make the MAC instructions and memory visiting coalesced and therefore improve the throughput. Under this setting, a SIMD *load* instruction is able to load $2^e$ values and a SIMD *MAC* instruction computes atleast $2^{e-1}$ values. Besides, the computer architecture society presents machine learning accelerators accelerating sparse matrix-vector multiplication in the sparse CERF. We continue to use the randomized autoencoder to show the effectiveness of group sparse CERF. The group size $2^e$ spans the range $\{1, 2, 4, 8, 16\}$. The other experimental settings are exactly the same with the previous subsection. We can see from Figure (1) and Table (5), the performance variance of this modification is negligible comparing to the improvement brought by CERF.

### 5. Conclusion

In this paper we propose a novel and generalized framework for training masked and blocked computation-efficient random features (CERF) from datasets. We also propose a novel approach of constrained variational expectation maximization which consists of ADMM optimizer during M-step. The experiments on various tasks and datasets demonstrate the significant improvement of CERF over RFF under any cost-budget, and the superiority to other kernel methods.